\documentclass[10pt,twocolumn]{IEEEtran}
\usepackage[utf8]{inputenc}
\usepackage[T1]{fontenc}
\usepackage{amsmath,amssymb,amsfonts}
\usepackage{graphicx}
\usepackage{booktabs}
\usepackage{tabularx}
\usepackage{array}
\usepackage{subcaption} 
\usepackage{float}
\usepackage{url}
\usepackage{cite}
\usepackage{xcolor}

\begin{document}


\title{Poster: Reliable 3D Reconstruction for Ad-hoc Edge Implementations\thanks{This  work supported by the National Science Foundation under Award Number: CNS-1943338 and OAC-2232889.}}

\thispagestyle{empty}

 \author{
Md. Nurul Absur$^\ast$, Swastik Brahma$^\$$, Saptarshi Debroy$^\ast$\\
$^\ast$City University of New York, $^\$$University of Cincinnati\\ Emails: \textit{mbsur@gradcenter.cuny.edu, brahmask@ucmail.uc.edu, saptarshi.debroy@hunter.cuny.edu}}
\maketitle

\begin{abstract}

Ad-hoc edge deployments to support real-time complex video processing applications such as, multi-view 3D reconstruction often suffer from spatio-temporal system disruptions that greatly impact reconstruction quality. In this poster paper, we present a novel portfolio theory-inspired edge resource management strategy to ensure reliable multi-view 3D reconstruction by accounting for possible system disruptions.

\end{abstract}


\section{Introduction}

Multi-view 3D reconstruction~\cite{10419312} applications are increasingly becoming fundamental for complex 3D video processing application pipelines that create immersive 3D environments, such as, virtual, augmented, and mixed reality. 
In order to provide rapid situational awareness, mission-critical use cases often rely on ad-hoc edge environments where: i) raw video data is captured by end-devices, such as, drones and robots; ii) speciality on-premise (e.g., hosted on vehicles) edge servers equipped with CPU/GPU and 3D video processing applications are deployed on-demand computation; and iii) ground consumers (e.g., tactical or first responder units) visualize the reconstructed scene on their hand-held devices.

The already tricky proposition of resource management in ad-hoc edge environments becomes even more challenging as such ad-hoc system deployments with weakly-coupled components are often prone to `disruptions' that can have potentially catastrophic impact the reconstruction quality. In general, such disruptions may occur due to various unintended factors, such as, abrupt changes in the operating environment, and/or malicious factors, such as jamming attacks on wireless channels.
These disruptions 
can lead to: i) `spatially correlated' failures in many cameras, all at the same time and/or ii) a series of `temporally correlated' failures across many cameras over a period of time. 
{\em Thus, there is a need for reliable camera management in the presence of such spatio-tempally correlated disruptions to ensure minimum reconstruction quality satisfaction. Although the related literature offers many solutions for edge resource management for critical use-cases~\cite{10.1145/3589639}, there exists little work on ensuring reliability in such environments in the presence of disruptions.}

In this paper, we propose an intelligent edge resource management technique that makes multi-view 3D reconstruction in such environments reliable under system-wide disruptions. 
In particular, our methodology involves an innovative camera selection strategy inspired by 
{\em portfolio theory} that is aimed at mitigating the potentially catastrophic effects of spatial and temporal disruptions that are commonplace in ad-hoc edge implementations~\cite{InvestmentScience-Luenberger,PortfolioSelection-Markowitz}. We evaluate the proposed edge resource management technique through a detailed simulation that mimics spatio-temporal correlated disruptions on ad-hoc edge implementations. Using publicly available and custom-made multi-view 3D reconstruction datasets, and state-of-the-art 3D reconstruction pipeline, we evaluate the quality of 3D reconstruction: a) with disruptions and addressed using traditional baseline approaches towards ensuring reliability and b) with disruption but resolved using our proposed portfolio theoretic approach. Our initial results demonstrate that under different degrees of system-wide disruptions and camera selection stipulations, our proposed method can ensure more reliable reconstruction in terms of higher mean value and lower standard deviation of reconstruction quality than baseline strategies.

\section{System Model and Problem Formulation}

\begin{figure}[t]
  \centering
  \includegraphics[width=0.43\textwidth]{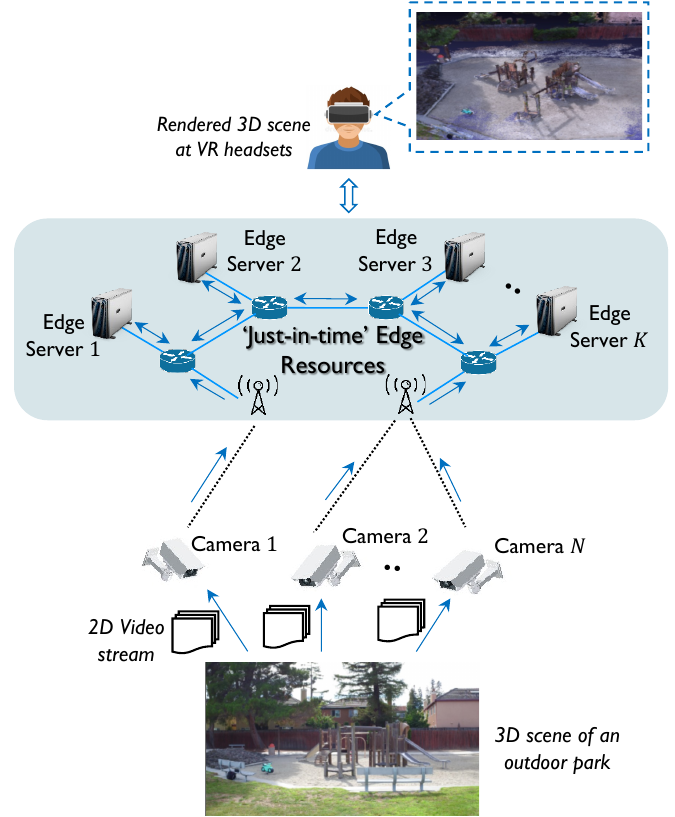}
  \caption{\footnotesize{Multi-view 3D reconstruction of an outdoor scene hosted at `Just-in-time' edge environment}}
  \label{Fig:sys_model}
  \vspace{-0.2in}
\end{figure}


Our ad-hoc edge system model, as depicted in Fig.~\ref{Fig:sys_model}, consists of a collection of $N$ camera-enabled devices, labeled as $\{1, \cdots, N\}$. 
Each of these cameras capture an indoor or outdoor 3D scene (an outdoor park in this case) from different viewpoints in the form of a continuous stream of video frames. The video streams are transmitted to one or multiple edge servers ($\{1, \cdots, K\}$) in real-time where the video frames from different cameras belonging to a single time-stamp are fused together using a certain 3D reconstruction algorithm, in order to generate a 3D visualisation of the scene for that particular time-stamp. 
We assume that a camera $i$, transmits images with specified image resolution $R_i$ with a probability $p_{c_i}$, and fails, producing no usable data, with a probability $1-p_{c_i}$. The likelihood of disruptions, represented as a random variable $p_{c_i}$, reflects the variability and uncertainty in camera operation. 
Such operational disruptions across different cameras can be correlated, indicated by the correlation coefficients $\rho_{i,j}$ for cameras $i$ and $j$. 

In such a scenario, it can be noted that the resolution of the image obtained from camera $i$, say denoted as $\mathcal{R}_i$, is itself a random variable, with $\mathcal{R}_i=0$ (when camera $i$ gets disrupted), which occurs with probability $1-p_{c_i}$) 
and $\mathcal{R}_i=R_i$ (when camera $i$ is not disrupted), with $\mathbb{E}[\mathcal{R}_i] = R_i\mathbb{E}[p_{c_i}]$ (where $\mathbb{E}[\cdot]$ denotes expectation), which occurs with probability $p_{c_i}$).
Further, for cameras $i$ and $j$, with $p_{c_i}$ and $p_{c_j}$ being correlated random variables, (with $\rho_{i,j}$ being their correlation coefficient), 
$\mathcal{R}_i$ and $\mathcal{R}_j$ also become correlated in nature 
whose covariance is $\mathrm{cov}(\mathcal{R}_i,\mathcal{R}_j) = R_i R_j \sigma_{p_{c_i}} \sigma_{p_{c_j}} \rho_{i,j}$, where $\sigma_{p_{c_i}}$ is the standard deviation of $p_{c_i}$.
To 
perform camera selection in such a scenario, suppose that 
camera $i$ is selected with a probability $\alpha_i$, with $\boldsymbol{\alpha} = (\alpha_1,\cdots,\alpha_N)$ denoting the camera selection vector, leading to the \textit{expected} total resolution of the images obtained from the cameras to be $\sum_{i=1}^N \alpha_i \mathbb{E}[\mathcal{R}_i]$.

Portfolio theoretic resource management in the above problem scenario would enable camera selection under {\em simultaneous} consideration of {\em expected}  
image qualities (resolutions) 
obtained from the cameras (which impact the reconstructed scene's quality) in terms of their resolutions and the  covariances among the 
image qualities 
obtained from the cameras (which impact reliability considerations), thereby permitting us to achieve {\em optimal tradeoffs} between the quality and the reliability of a 3D reconstruction task in order to 
sustain its desired performance under disruptions.
\begin{subequations}
\label{cameraSelectOptProblem}
\begin{align}
\underset{\boldsymbol{\alpha}=(\alpha_1,\cdots,\alpha_N)}{\text{minimize}}
&\qquad \hspace{-0.25in} \hspace{-0.1in}~~\sum_{i = 1}^N \sum_{j=1}^N \alpha_i \alpha_j \mathrm{cov}(\mathcal{R}_i, \mathcal{R}_j) \label{eq:PortfolioThPrelimObj} \\
\text{Subj. to}
&\qquad  \sum_{i=1}^N \alpha_i \mathbb{E}[\mathcal{R}_i] \geq \Theta~~\text{(\textit{quality constr.})}\label{eq:qualityConstraint} \\
&\hspace{-0.27in}~~~~~~~~~~~\sum_{i=1}^N \alpha_i \leq \Psi~~\text{(\textit{resource constr.})} \label{eq:resourceConstraint} \\
&\qquad  \hspace{-0.27in}~~~~~~\alpha_i \in [0,1],~~\forall i~~\text{(\textit{sel. param. constr.})}  \label{eq:probabilityFeasibility}
\end{align}
\end{subequations}

It should be noted that Eq.~\eqref{cameraSelectOptProblem} seeks to determine
the camera selection strategy $\boldsymbol{\alpha} = (\alpha_1,\cdots,\alpha_N)$ that would, to provide \textit{reliability}, minimize the quantity $\sum_{i = 1}^N \sum_{j=1}^N \alpha_i \alpha_j \mathrm{cov}(\mathcal{R}_i, \mathcal{R}_j)$, which determines the \textit{variability} of the quality of the reconstructed scene. For this work, we use genetic algorithm to solve Eq.~\eqref{cameraSelectOptProblem}.

\section{Experiments and Initial Results}
\begin{figure}[t]
    \centering
    \begin{subfigure}[b]{0.48\columnwidth} 
        \centering
        \includegraphics[width=\textwidth]{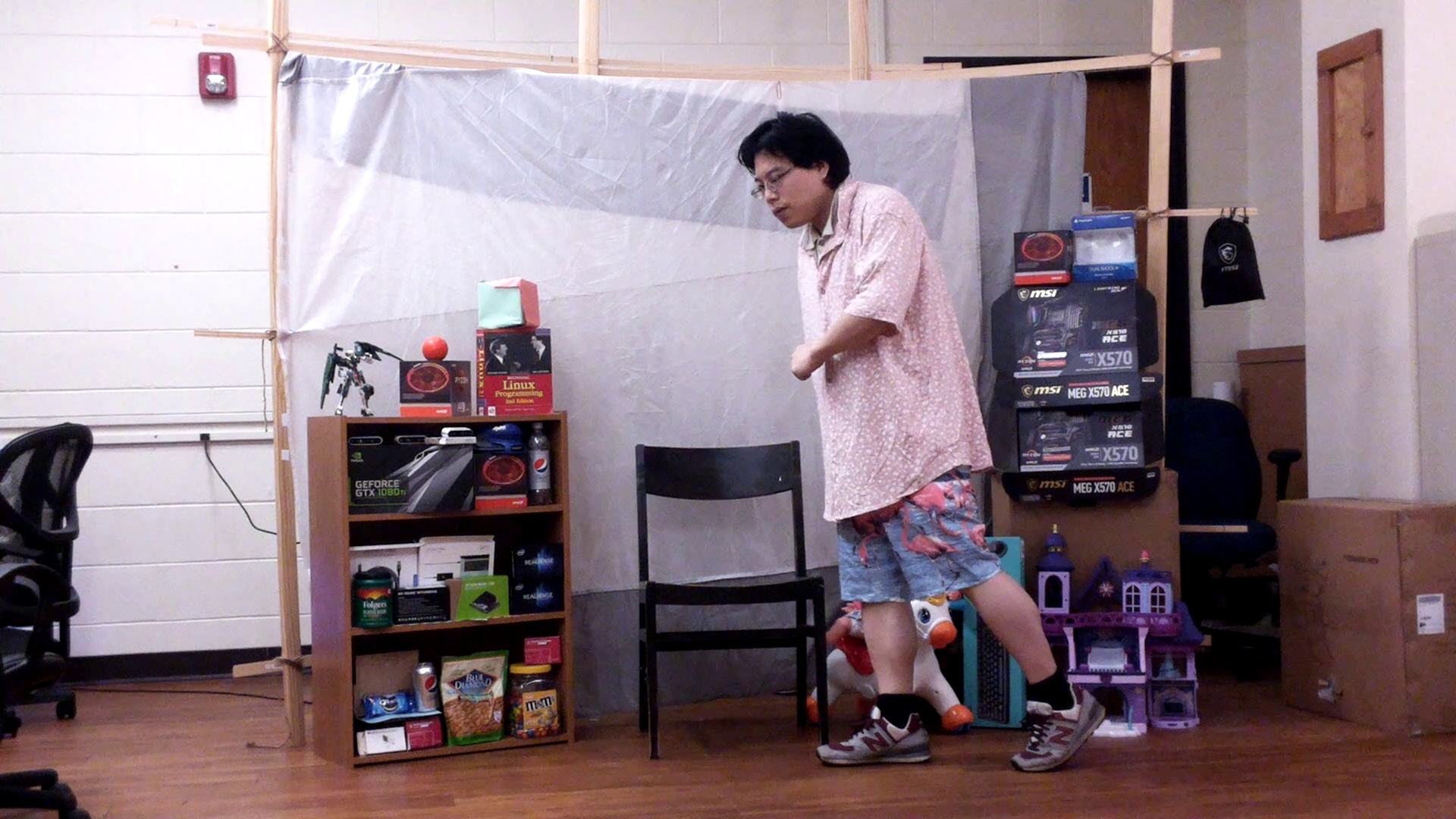}
        \caption{\em Walk}
        \label{fig:walk}
    \end{subfigure}
    \hfill 
    \begin{subfigure}[b]{0.48\columnwidth}
        \centering
        \includegraphics[width=\textwidth]{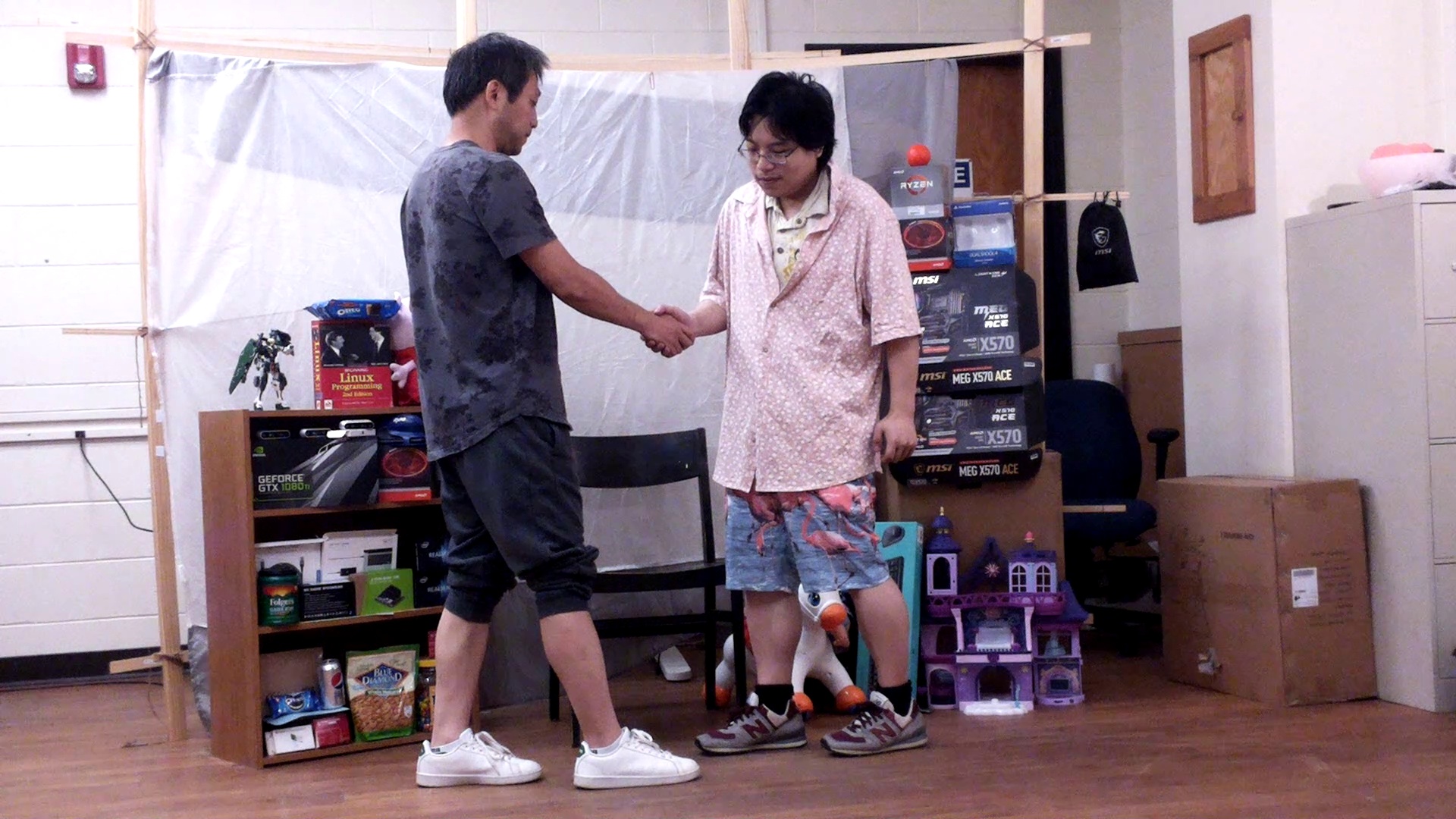}
        \caption{\em Handshake}
        \label{fig:handshake}
    \end{subfigure}

    \caption{\footnotesize Image samples from the captured multi-view dataset.}
    \label{fig:Sample}
    \vspace{-0.2in}
\end{figure}

For the evaluation, we use the most widely used multiview 3D reconstruction algorithm, viz., openMVG/openMVS~\cite{moulon2016openmvg,openmvs2020}, 
running on a Dell desktop with Intel i7 @2.9GHz processor with 16GB RAM, and NVIDIA GeForce RTX 2060.
For the dataset, we use publicly available Dance1 and Odzemok~\cite{MustafaICCV15} dataset (7 camera video stream), as well as generate our own 5 camera video stream multi-view datasets representing different degrees of indoor dynamic scenes, viz., {\em Walk} and {\em Handshake} as shown in 
Fig.\ref{fig:Sample}. 

For the experiments, we create a heterogeneous environment where different cameras have different image resolution, i.e., cameras with higher resolution are more important in generating high quality reconstruction. We further simulate disruption events which cause the cameras to not send their data for reconstruction uniformly randomly, however, with pairwise disruption being correlated with beta distribution. Given such environment, we generate comparison results over multiple runs simulating long periods of ad-hoc edge environmental operation and multiple occurrence of potential disruption events.  

\begin{figure}[t]
    \centering
    \includegraphics[width=\columnwidth]{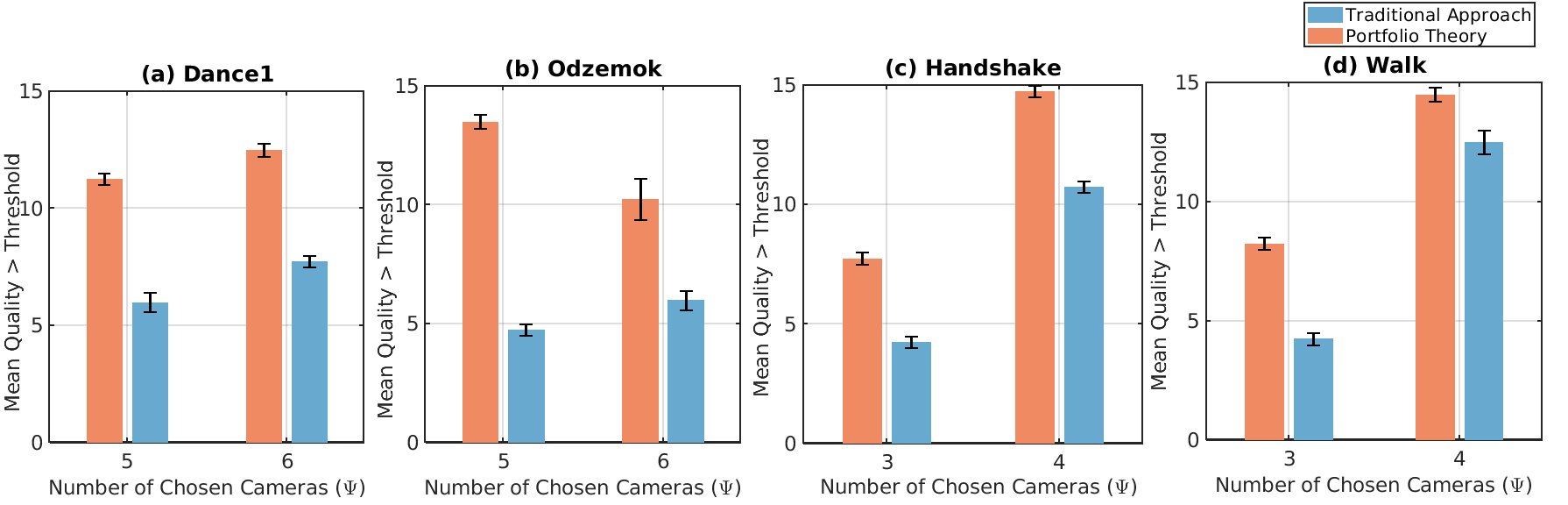}
    \caption{\footnotesize Reliability of reconstruction quality comparison between portfolio theoretic vs. baseline approaches.}
    \label{Fig:evidence}
\end{figure}

Fig.~\ref{Fig:evidence} demonstrates our initial results where we compare the reliability of reconstruction quality between two competing camera selection strategies under system-wide disruptions: a) our proposed portfolio theory based and b) traditional mean expected quality based where cameras with higher resolution are selected more often. The results demonstrate that for different values of the number of chosen cameras (i.e., $\psi$), the mean reconstruction quality (over multiple experiments) using our proposed portfolio theoretic approach significantly outperforms reconstruction with the baseline strategy in terms of reliablity, i.e., generating useful 3D reconstructions with quality greater than a pre-defined threshold, selected through empirical analysis.


\section{Conclusions}

In this poster paper, we proposed a novel portfolio-theoretic approach to address potential disruptions in ad-hoc edge environments that can impact quality of underlying 3D reconstruction applications. Our initial results with real-world data and ad-hoc edge testbed demonstrate our proposed method's utility in improving reliability.


\end{document}